%% file: main.tex
\documentclass[acmtog, nonacm]{acmart}

\acmSubmissionID{2280}

\usepackage{booktabs} % For formal tables

\usepackage[ruled]{algorithm2e} % For algorithms

\SetAlFnt{\small}
\SetAlCapFnt{\small}
\SetAlCapNameFnt{\small}
\SetAlCapHSkip{0pt}

\input{preamble}

\begin{document}
% Title portion
\title{RealCam: Real-Time Novel-View Video Generation with Interactive Camera Control}

% DO NOT ENTER AUTHOR INFORMATION FOR ANONYMOUS TECHNICAL PAPER SUBMISSIONS TO SIGGRAPH 2019!
\author{Youcan Xu}
\authornote{Work was done during an internship at Xmax.AI.}
\orcid{0009-0001-3386-5014}
\affiliation{%
 \institution{Zhejiang University}
 \city{Hangzhou}
 \country{China}}
\email{youcan@zju.edu.cn}

\author{Jiaxin Shi}
\affiliation{%
 \institution{Xmax.AI Ltd.}
 \city{Beijing}
 \country{China}
}
\email{jiaxin@xmax.ai}

\author{Zhen Wang}
\affiliation{%
\institution{Hong Kong University of Science and Technology}
\city{Hong Kong}
\country{China}}
\email{zhenwang@ust.hk}

\author{Wensong Song}
\affiliation{%
 \institution{Zhejiang University}
 \city{Hangzhou}
 \country{China}
}
\email{wensong.song@zju.edu.cn}

\author{Feifei Shao}
\affiliation{%
 \institution{Zhejiang University}
 \city{Hangzhou}
 \country{China}}
\email{sff@zju.edu.cn}

\author{Chen Liang}
\affiliation{%
 \institution{Xmax.AI Ltd.}
 % \city{Guangzhou}
 \country{China}
}
\email{chenliang2@hkust-gz.edu.cn}

\author{Jun Xiao}
\affiliation{%
 \institution{Zhejiang University}
 \city{Hangzhou}
 \country{China}
}
\email{junx@cs.zju.edu.cn}

\author{Long Chen}
\authornote{Long is the corresponding author.}
\affiliation{%
 \institution{Hong Kong University of Science and Technology}
 \city{Hong Kong}
 \country{China}
}
\email{longchen@ust.hk}

\begin{abstract}
Camera-controlled video-to-video (V2V) generation enables dynamic viewpoint synthesis from monocular footage, holding immense potential for interactive filmmaking and live broadcasting. However, existing implicit synthesis methods fundamentally rely on non-causal, full-sequence processing and rigid prefix-style temporal concatenation. This architectural paradigm mandates bidirectional attention, resulting in prohibitive computational latency, quadratic complexity scaling, and inherent incompatibility with real-time streaming or variable-length inputs. To overcome these limitations, we introduce \texttt{RealCam}, a novel autoregressive framework for interactive, real-time camera-controlled V2V generation. We first design a high-fidelity teacher model grounded in a \textbf{Cross-frame In-context Learning} paradigm. By interleaving source and target frames into synchronized contextual pairs, our design inherently enables length-agnostic generalization and naturally facilitates causal adaptation, breaking the rigid prefix bottleneck. We then distill this teacher into a few-step causal student via Self-Forcing with Distribution Matching Distillation, enabling efficient, on-the-fly streaming synthesis. Furthermore, to mitigate severe loop inconsistency in closed-loop trajectories, we propose \textbf{Loop-Closed Data Augmentation (LoopAug)}, a novel paradigm that synthesizes globally consistent loop sequences from existing multiview datasets. Extensive experiments demonstrate that \texttt{RealCam} achieves state-of-the-art visual fidelity and temporal consistency while enabling truly interactive camera control with orders-of-magnitude faster inference than existing paradigms. Our project page is at \url{https://xyc-fly.github.io/RealCam/}.
\end{abstract}




% \keywords{Wireless sensor networks, media access control,
% multi-channel, radio interference, time synchronization}

\maketitle

\input{tex/1-intro}

\input{tex/2-related_works}
\input{tex/3-method}

\input{tex/4-experiments}
\input{tex/5-conclusion}
%%reference
\bibliographystyle{ACM-Reference-Format}
\bibliography{sample-bibliography}

\end{document}

%% file: preamble.tex
\usepackage{color}
\definecolor{citecolor}{RGB}{66,168,235}
\definecolor{linkcolor}{RGB}{255,0,0}
\hypersetup{colorlinks=true,citecolor=citecolor,linkcolor=linkcolor}
\usepackage{xcolor}         % colors

\usepackage{pifont}% http://ctan.org/pkg/pifont
\definecolor{tabfirst}{rgb}{1, 0.7, 0.7}
\definecolor{tabsecond}{rgb}{1, 0.85, 0.7}
\definecolor{tabthird}{rgb}{1, 1, 0.7}
\usepackage{makecell}
\usepackage{animate}
\usepackage{graphicx}	
\usepackage{amsmath}	
\usepackage{mathtools}
\usepackage{amsthm}
\usepackage{booktabs}
\usepackage{times}
\usepackage{epsfig}
\usepackage{caption}
\usepackage{float}
\usepackage{placeins}
\usepackage{color, colortbl}
\usepackage{enumitem}
\usepackage{tabularx}
\usepackage{xstring}
\usepackage{multirow}
\usepackage{xspace}
\usepackage{url}
\usepackage{subcaption}
\usepackage[hang,flushmargin]{footmisc}
\usepackage{kotex}
\usepackage{arydshln}
\usepackage{adjustbox}
\usepackage{wrapfig}
\usepackage{algorithmicx,algpseudocode}
\usepackage{listings}
\usepackage{bm}

\newlength\paramargin
\newlength\abovetabcapmargin
\newlength\belowtabcapmargin
\newlength\abovefigcapmargin
\newlength\belowfigcapmargin
\newlength\aboveeqmargin
\newlength\beloweqmargin

\usepackage{thmtools,thm-restate}

\newcommand{\eg}{\textit{e}.\textit{g}.}
\newcommand{\ie}{\textit{i}.\textit{e}.}
\newcommand{\cf}{\textit{cf.}}

%% file: tex/1-intro.tex
\section{Introduction}
Camera movement serves as a fundamental cinematic language that directs audience attention and evokes profound emotional resonance. However, achieving professional-level camera movement such as sweeping crane shots or cinematic dolly-ins remains a privilege of high-budget productions. This typically requires expensive hardware rigs, meticulous pre-production planning, and costly reshoots, which places advanced cinematography out of reach for amateur videographers and content creators. To address this dilemma, recent advances in generative AI, especially the camera-controlled video-to-video (V2V) generative model~\cite{vanhoorick2024gcd,ICCV_recammaster,CVPR_redirector, luo2025camclonemaster,ICCV_TrajectoryCrafter,NIPS_cognvs}, have emerged as a transformative post-production paradigm. By synthesizing novel viewpoints directly from existing footage, this paradigm allows for the post-hoc synthesis of user-specified viewpoints from casual captures, significantly reducing production costs and democratizing high-quality cinematography~\cite{ICCV_recammaster,luo2025camclonemaster}.

\begin{figure}[t]
    \centering
    \includegraphics[width=0.95\linewidth]{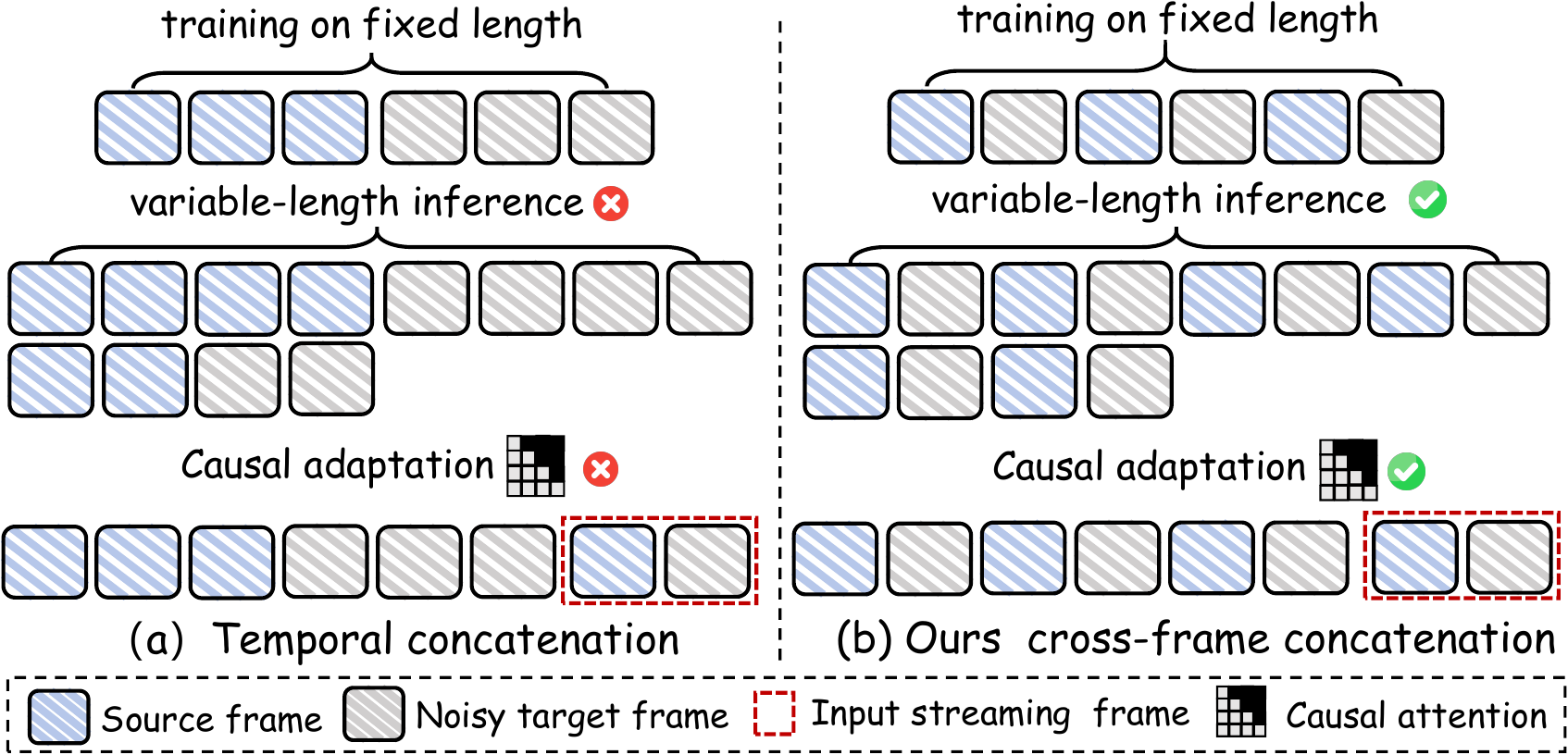}
    \caption{ \textbf{Comparison of direct temporal concatenation and ours cross-frame concatenation.} Our method generalizes to arbitrary video length during inference and naturally extends to causal attention.}
    \label{fig:token_merge_strategy}
\end{figure}

To achieve this goal, existing approaches can be broadly categorized into two groups.
One prominent line of research employs warp-then-inpaint frameworks~\cite{Traj_attention,ICCV_TrajectoryCrafter, ReCapture, jeong2025reangle, lu2025see4d,NIPS_cognvs}. These approaches utilize estimated depth~\cite{hu2025-DepthCrafter} to perform explicit geometric transformations, getting geometrically aligned but content-missing warped frames. Then, a generative model conditioning on these frames is employed to refine and inpaint them. However, these methods are sensitive to per-frame depth accuracy and generate degraded results under large viewpoint changes and complex scene structure.
Alternatively, implicit synthesis methods~\cite{vanhoorick2024gcd,ICCV_recammaster,CVPR_redirector} bypass explicit warping by directly injecting camera representations into the latent space to learn multi-view relationships. A representative method is ReCamMaster~\cite{ICCV_recammaster}, which harnesses the generative power of pre-trained text-to-video models~\cite{wang2025wan} through a dedicated video conditioning mechanism. Specifically, it concatenates source and target video tokens along the frame dimension and injects {encoded camera parameters} into the backbone's feature stream. 
During fine-tuning on large-scale synchronized datasets~\cite{ICCV_recammaster}, the 3D full attention of the base model naturally aggregates information across both sequences, enabling the model to implicitly internalize cross-view geometric correspondences without relying on external depth estimator models. 

% \begin{wrapfigure}[15]{t}{0.6\textwidth}
%     \vspace{-1.5em}
%     \centering
%     \includegraphics[width=0.95\linewidth]{figures/token_strategy_vs_ recam_1.pdf}
%     \caption{ \textbf{Comparison of direct temporal concatenation and ours cross-frame concatenation.} Our method generalizes to arbitrary video length during inference and naturally extends to causal attention.}
%     \label{fig:token_merge_strategy}
% \end{wrapfigure}

Despite achieving impressive visual fidelity, existing methods from both categories are fundamentally precluded from real-time, interactive applications due to prohibitive computational latency.
This bottleneck is rooted in their reliance on bidirectional attention mechanisms --- an inherently non-causal design that mandates all control inputs be specified a priori, trapping users in a frustrating ``render-and-wait'' cycle~\cite{shin2025motionstream}. For instance, state-of-the-art models like ReCamMaster~\cite{ICCV_recammaster} take over 17 minutes to synthesize a mere 5-second clip. 

Furthermore, we identify another fundamental structural barrier in prevailing implicit synthesis frameworks~\cite{ICCV_recammaster,CVPR_redirector}:
The use of temporal concatenation. As illustrated in Figure~\ref{fig:token_merge_strategy}(a), current SOTA models typically append the target video tokens after the complete source video sequence. This rigid prefix-style conditioning introduces two critical limitations: 
\vspace{-0.8em}
\begin{itemize}[leftmargin=*]
\itemsep-0.2em

\item \textbf{Generalization Collapse on Variable Lengths.} The prefix-style concatenation~\cite{ICCV_recammaster,luo2025camclonemaster} inherently binds the model to fixed-length training regimes. Consequently, it fails to generalize to variable-length inputs during inference, exhibiting severe fidelity degradation and temporal incoherence when sequence lengths deviate from the training distribution.

\item \textbf{Architectural Incompatibility with Streaming.} The rigid prefix dependency forces the model to encode the entire source context, even before generating the first target frame. 
Such a strictly non-causal design fundamentally precludes frame-by-frame generation~\cite{yin2025Causvid,self_forcing},
making it architecturally incompatible with streaming or interactive workflows (\eg, interactive live streaming), where dynamic, {on-the-fly} viewpoint manipulation is essential for sustaining audience engagement.
\end{itemize}
\vspace{-0.5em}

Despite recent efforts~\cite{CVPR_redirector} that incorporate camera-conditioned RoPE~\cite{rope} to support variable lengths still retain this rigid prefix structure and suffer from quadratic computational scaling. Consequently, achieving scalable, interactive generation remains fundamentally intractable under existing paradigms. 

To address these challenges, we propose \texttt{RealCam}, the first framework capable of real-time, interactive camera-controlled V2V generation without explicit wrapping. Built upon a state-of-the-art DiT-based model~\cite{wang2025wan}, \texttt{RealCam} adopts a two-phase design that directly addresses the aforementioned bottlenecks: \textbf{1)  Bidirectional Teacher Training via Cross-frame In-context Learning}  to construct a length-agnostic, causal-ready teacher model, and \textbf{2) Causal Distillation with LoopAug} for efficient streaming inference with global consistency. In the first phase,  our core insight is to refactor the video-to-video relationship by interleaving source and target frames into unified contextual pairs, enabling the model to generalize to arbitrary video lengths and naturally transition to causal attention (\cf, Figure~\ref{fig:token_merge_strategy}(b)). In the second phase, building upon this cross-frame teacher, we distill it into a causal student through self-forcing rollout, augmented with our proposed Loop-Closed Data Augmentation (LoopAug) to resolve temporal drift in cyclic trajectories.

\textbf{In the Bidirectional Teacher Training phase}, we employ a structural shift from prefix-style conditioning to synchronous frame-pair processing. Instead of treating the source video as a long-range prefix, we interleave source and target frames into a unified sequence of contextual pairs. This cross-frame design provides two decisive advantages: (i) \textit{Causal Compatibility:} By processing each frame-pair synchronously, the teacher model can naturally transition to a causal adaptation (\ie, replacing bidirectional attention with causal masks) without breaking the conditional dependency, enabling frame-by-frame generation. (ii) \textit{Dynamic Robustness:} Our cross-frame paradigm focuses on relative positional relationships rather than absolute sequence length, allowing the model to generalize to arbitrary video lengths during inference. \textbf{In the Causal Distillation phase}, we distill the cross-frame teacher into a few-step causal student  {through self-forcing-style distillation~\cite{self_forcing, zhao2025real_motion} for efficient, low-latency inference}. However, we identify a critical limitation in direct distillation: severe loop inconsistency arises in closed-loop trajectories, causing visual distortions when the camera returns to its starting viewpoint. We attribute the challenge to a lack of loop-consistent supervision during distillation. To address this without costly manual data collection, we introduce \emph{Loop-Closed Data Augmentation (LoopAug)}, a novel paradigm that synthesizes loop-closed sequences from existing multiview datasets, enabling long-horizon video generation with global consistency.

% The contributions of this work can be summarized as follows. 1) We propose RealCam, an efficient, real-time framework for interactive, camera-controlled video-to-video generation. 2) We design a cross-frame in-context learning mechanism for the teacher model, enabling inherent causal compatibility and length-agnostic generalization. 3) We propose LoopAug, a simple yet effective data augmentation paradigm that significantly improves loop consistency and global stability in long-horizon video generation. 4) Extensive experiments and ablation studies demonstrate that RealCam achieves superior performance and efficiency compared with existing state-of-the-art methods.

% itemize
The contributions of this work can be summarized as follows. 
\begin{itemize}
\item To the best of our knowledge,  RealCam is the first
real-time framework for interactive, camera-controlled video-to-video generation without explicit wrapping~\cite{ICCV_TrajectoryCrafter, lu2025see4d, Traj_attention}.
\item We design a cross-frame in-context learning mechanism for the teacher model, enabling inherent causal compatibility and length-agnostic generalization.
\item We propose LoopAug, a simple yet effective data augmentation paradigm that significantly improves loop consistency and global stability in long-horizon video generation.
\item 
Extensive experiments and ablation studies demonstrate that RealCam achieves superior performance and efficiency compared with existing state-of-the-art methods.
\end{itemize}

%% file: tex/2-related_works.tex
\section{Related Work}
\label{sec:related_works}
\noindent\textbf{Camera-Controlled Video Generation.} With the goal of steering viewpoint changes through camera conditions,
camera-controlled video generation has become an important branch of controllable video synthesis~\cite{ma2025controllable_survey}. Early studies primarily explored this capability in text-to-video~\cite{he2024cameractrl, wang2024motionctrl, kuang2024collaborative, AC3D, wang2025cinemaster, hou2024training, ling2024motionclone, zhao2024motiondirector} or image-to-video~\cite{viewcraft, Li_2025_ICCV, feng2024i2vcontrol,xu2024camco, Traj_attention} settings. More recently, growing attention has shifted toward camera-controlled video-to-video generation~\cite{vanhoorick2024gcd,ICCV_recammaster, luo2025camclonemaster}. This challenging task necessitates maintaining rigorous temporal synchronization and semantic consistency with the source video while simultaneously re-projecting it onto a user-provided camera trajectory. 
Current methodologies generally follow two technical routes. Warp-then-inpaint methods \cite{Traj_attention, ICCV_TrajectoryCrafter, lu2025see4d, ReCapture, jeong2025reangle} typically employ depth estimation to lift 2D frames into 3D point clouds, from which geometrically aligned proxy videos are rendered from target viewpoints. These proxies then serve as auxiliary conditions to guide the generative process.  Conversely, implicit synthesis methods \cite{ICCV_recammaster,CVPR_redirector} directly encode camera parameters into the model, often leveraging large-scale synthetic datasets for training. Despite their impressive visual quality, both paradigms are predominantly designed for offline, full-sequence modeling, rendering them computationally prohibitive for real-time interactive control.

\noindent\textbf{Autoregressive Video Diffusion Models.}
Traditional video diffusion models~\cite{yang2024cogvideox, wang2025wan, kong2024hunyuanvideo} typically rely on full-sequence, bidirectional modeling, which limits their scalability for long-horizon generation and real-time streaming. To overcome these constraints, a growing line of research~\cite{chen2024diffusion_forcing, Ca2-VDM, Pyramidal, yin2025Causvid} integrates autoregressive (AR) prediction with diffusion modeling. Early attempts like MAGI-1~\cite{teng2025magi} and CausVid~\cite{yin2025Causvid} transition from bidirectional to causal architectures, enabling chunk-wise rollout.
To mitigate the inherent challenges of temporal drift and error accumulation in AR sampling, recent works have introduced advanced training and inference strategies. For instance, Self-Forcing~\cite{self_forcing} and Rolling Forcing~\cite{Rolling_forcing} address the train-test mismatch by conditioning the model on its own generated outputs or employing joint denoising schemes. Furthermore, Deep Forcing~\cite{yi2025deep} and LongLive~\cite{yang2025longlive} leverage attention-sink mechanisms and KV-cache management to maintain global consistency during extended rollouts. However, these methods are primarily optimized for static, prompt-based generation. While effective for visual quality, they often lack the flexibility required for fine-grained, user-driven control or real-time interactive adjustments.

\noindent\textbf{Interactive Video Models.} 
Interactive video models~\cite{Hunyuan-gamecraft, mao2025yume, zhao2025real_motion, shin2025motionstream, parker2024genie} aim to simulate responsive environments that support real-time user engagement and feedback. Recent advances have explored various interactive modalities to steer video content, ranging from motion trajectories~\cite{zhao2025real_motion, shin2025motionstream} and facial keypoints~\cite{li2025personalive} to multimodal signals like audio~\cite{su2026omniforcing, chern2025livetalk} and instructions~\cite{yesiltepe2025infinity, li2025egoedit}. 
While promising, these models typically excel at semantic or object-level manipulations but struggle with the geometric precision required for fluid, real-time viewpoint transitions in open-domain videos. 
In this work, we bridge this gap by focusing on generating novel views of input videos through interactive camera control, enabling an immersive "walk-through" experience with sub-second latency.

% For instance, \textbf{MotionStream}~\cite{} and \textbf{AR-Drag}~\cite{} enable users to manipulate object motion via interactive "drags" or trajectories, while game-driven models like \textbf{Hunyuan-GameCraft}~\cite{} and \textbf{Genie}~\cite{} unify discrete actions into continuous latent spaces to simulate gameplay. 

%% file: tex/3-method.tex
\section{Method}
\label{method}

\begin{figure*}[t]\centering
\includegraphics[width=0.95\textwidth]{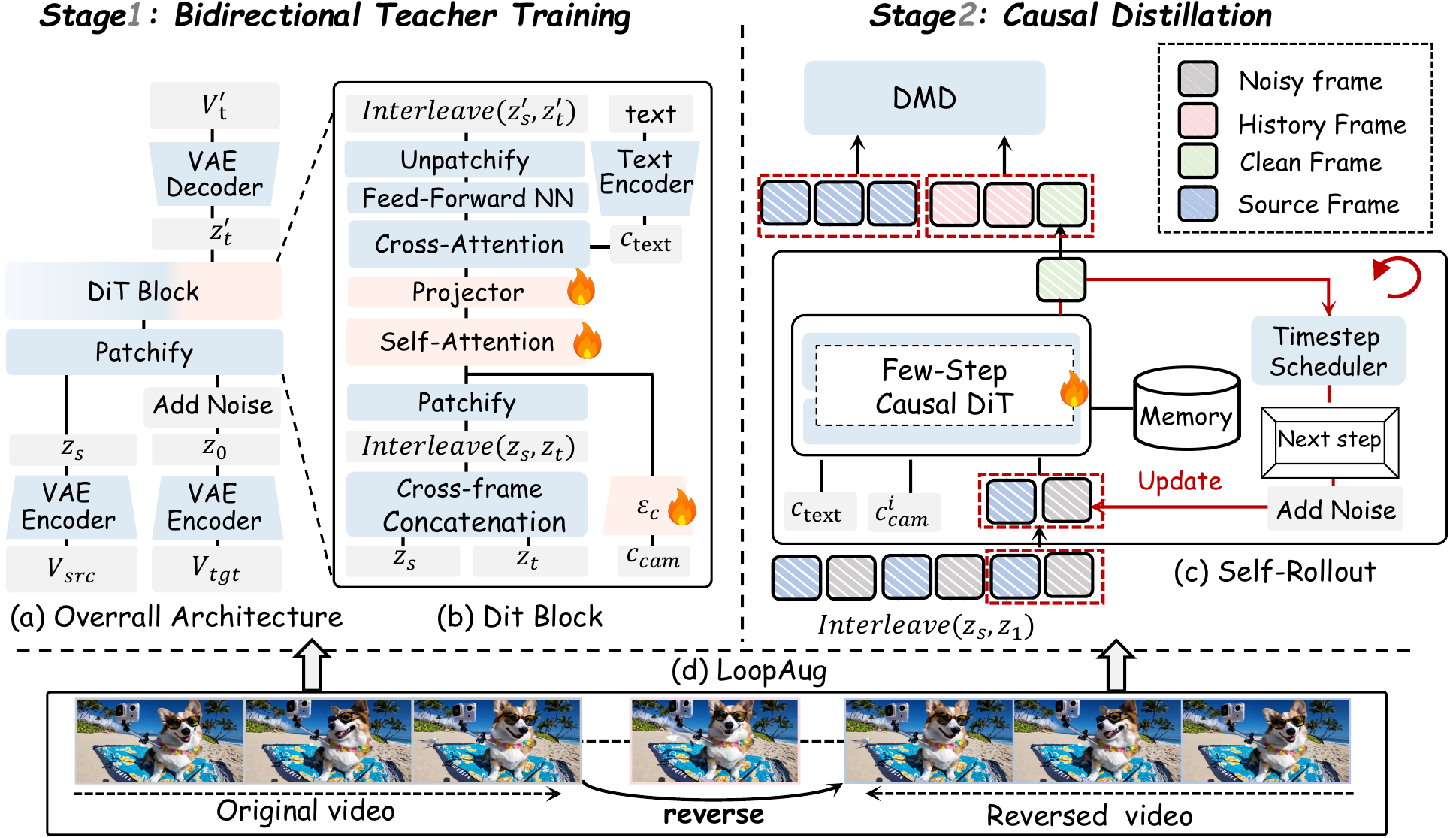}
\caption{\textbf{Model architecture and training pipeline.} \textit{Left:} The training pipeline of  teacher camera-controlled video-to-video model. (a) A latent diffusion model is optimized to reconstruct the target video $V_{tgt}$, conditioned on the source video $V_{src}$, target camera pose $c_{\text{cam}}$, and target prompt $c_{\text{text}}$. (b) We propose Cross-frame Concatenation to inject the source video condition, and following  ReCamMaster~\cite{ICCV_recammaster}, we train only self-attention layers, the camera encoder and projector.
\textit{Right:} Distilling a few-step causal diffusion model through Self Forcing-style DMD distillation. (c) The autoregressive self-rollout process with KV cache (memory) for generating a frame during both training and inference. After rolling out the full video sequence, the model is optimized by the DMD loss.
The bottom is the Loop-Closed Data Augmentation (LoopAug) strategy.}
    \label{fig_pipe}
\vspace{-1.0em}
\end{figure*}

Our \texttt{RealCam} aims to achieve real-time, interactive camera-controlled video-to-video (V2V) generation. To this end, we propose a two-step training pipeline. In Sec .~\ref{train_teacher}, we first train a bidirectional camera-controlled V2V model that leverages cross-frame in-context learning to achieve high-fidelity novel-view synthesis with robust length generalization. Then in Sec. ~\ref{train_student}, we distill this slow teacher model into a few-step c causal student model designed for streaming inference. To address the visual drift in long-horizon and closed-loop trajectories, we introduce LoopAug, a data augmentation strategy that enforces global consistency without additional manual labeling. The overview of the pipeline is depicted in Figure ~\ref{fig_pipe}.
\subsection{Preliminaries }
\label{sec:pre}
\textbf{Text-to-Video Base Model.}
Our work builds upon the Wan DiT family~\cite{wang2025wan}, a latent video diffusion model that performs generative modeling directly in the latent space.  This model comprises three key components: a 3D Variational Auto-Encoder (VAE),
a Transformer-based velocity prediction network, and a text encoder.
Given an input video  $x  \in \mathbb{R}^{F\times C \times H \times W}$, the VAE encoder $\mathcal{E}$  compresses it into the latent space  $z_0 = \mathcal{E}(x)  \in \mathbb{R}^{f\times c \times h \times w}$. The generative process aims to model the conditional distribution $p(z_0 | c_{\text{text}})$, where $c_{\text{text}}$ denotes the textual embedding derived from the text encoder.
We adopt the Flow Matching (FM)~\cite{flux2024,esser2024scaling,wang2025wan} for generative modeling,  where the FM learns a transformation from a standard Gaussian distribution $z_1 \sim \mathcal{N}(0, \mathbf{I})$ to the target $z_0$, and decodes it back to clean data with the decoder $x = \mathcal{D}(z_0)$.
Specifically, for each training step with timestep $t\sim \left [ 0, 1 \right ]$, FM obtains a noised intermediate latent via the linear interpolation between $z_0$ and $z_1$ as $z_t = (1-t) z_0 + t z_1 $. The velocity prediction network $v_\theta$ is then trained to estimate the velocity $v_t = \frac{dz_t}{dt} = z_1 - z_0$ by minimizing the flow matching loss:
\begin{equation}
\label{eq:fm_training_objective}
\mathcal{L}_\mathrm{FM}=\mathbb{E}_{z_0,z_1,t,c_{text}}\left[\|v_\theta(z_t,t, c_{text})- (z_1 - z_0)\|_2^2 \right].
\end{equation}
During inference, the FM starts from a random noisy latent $z_1 \sim \mathcal{N}(0, \mathbf{I})$, and progressively integrates the predicted velocity to obtain the clean latent: $z_0 = z_1 - \int_{1}^{0}v_\theta(z_t,t, c_{\text{text} })dt$.

\textbf{Distribution Matching Distillation .} Distribution matching distillation (DMD)~\cite{dmd,improved_dmd} is a technique that distills multi-step teacher diffusion model into a few-step student model. The core idea is to minimize KL divergence between the real data distribution $p_{t}^{\text{data}}$ (approximated by the frozen teacher model) and the student-generated distribution $p_{\text{t}}^{\text{gen}}$ across randomly sampled time $t$:  $\mathcal{L}_{\text{DMD}} = \mathbb{E}_t \left[ D_{\text{KL}}(p_{t}^\text{gen} \| p_{t}^{\text{data}}) \right]$. The loss is optimized by descending along its gradient:
% \begin{equation}
% \label{dmd_score}
%     \nabla_\theta \mathcal{L}_{\text{DMD}} \approx -\mathbb{E}_{t} \left[ \left( s_{\text{real}}(\Psi(G_{\theta}(\epsilon),t), t) - s_{\text{fake}}(\Psi(G_{\theta}(\epsilon),t), t) \right) \cdot \frac{d G_{\theta}(\epsilon)}{d \theta}d\epsilon \right],
% \end{equation} 

\begin{equation}
\label{dmd_score}
\begin{split}
    \nabla_\theta \mathcal{L}_{\text{DMD}} \approx -\mathbb{E}_{t} \biggl[ & \Bigl( s_{\text{real}}(\Psi(G_{\theta}(\epsilon),t), t) \\
    & - s_{\text{fake}}(\Psi(G_{\theta}(\epsilon),t), t) \Bigr) \cdot \frac{d G_{\theta}(\epsilon)}{d \theta} d\epsilon \biggr ].
\end{split}
\end{equation}

where $\Psi$ is the forward diffusion process, 
$\epsilon$ is random Gaussian noise, $G_{\theta}$ is the generator parameterized by $\theta$,
$s_{\text{real}}$ is the frozen score function for real data while $s_{\text{fake}}$  is the learnable score function trained on the generator’s outputs. During training, DMD initializes both score functions from the teacher model.

\subsection{Bidirectional Teacher Model Training}
\label{train_teacher}
In the Bidirectional Teacher Training phase, we aim to build a bidirectional teacher model, which conditions on an fixed length input video $V_{src} \in \mathbb{R}^{F\times C \times H \times W}$  and generates a re-rendering video $V_{tgt} \in \mathbb{R}^{F\times C \times H \times W}$ that follows the user-specified trajectories. The target trajectories is represented as the extrinsic 
parameters of the camera denoted by $\texttt{c}_{cam} \coloneqq [{R}^{wc}, {t}^{wc}] \in \mathbb{R}^{F\times3\times4}$.
We start from the pretrained  text-to-video generative model \cite{wang2025wan}
and finetune it into a video-to-video generative model.

% A key challenge in camera-controlled V2V generation is maintaining synchronization and content consistency with the source video. 
% Existing methods~\cite{ICCV_recammaster, luo2025camclonemaster} directly concatenate the source video frames with the target video frames along the temporal dimension.
% However, as shown in Figure~\ref{fig:token_merge_strategy}, this strategy 
% treats absolute fixed-length source videos as context conditions for training,  which limits their generalization to variable-length sequences during inference.  Besides, this rigid concatenation is incompatible with the following causal adaptation described in Sec.~\ref{casual_adaption}. 
% forces the model to process fixed-length source video as  input videos and is incompatible with the following causal adaptation described in Sec.~\ref{casual_adaption}. 
% To address this challenge, we propose Cross-frame In-context condition injection. Instead of simple concatenation, we treat the source video  frames as the relative position condition:
\textbf{Cross-frame In-context Learning.}
To maintain dynamic synchronization and content consistency with the source video, while without relying on fixed-length temporal prefixes. We employ a cross-frame conditioning scheme. Rather than concatenating source and target latents along the temporal axis, we interleave them at the frame level:
\begin{equation}
\label{cross_frame}
\left\{
\begin{aligned}
    &z_{s} = \mathcal{E}(V_{src})  , \quad z_0 = \mathcal{E}(V_{tgt}), \\ 
    & \texttt{Interleave}(z_{s}, z_t)= [z_{s}^{1},z_{t}^{1},z_{s}^{2},z_{t}^{2},\dots z_{s}^{f},z_{t}^{f}]  ,
\end{aligned}
\right.
\end{equation}

% \begin{wrapfigure}[21]{r}{0.6\textwidth}
%     \vspace{-1.em}
%     \centering
%     \includegraphics[width=0.95\linewidth]{figures/recam-ous.pdf}
%     \vspace{-.6em}
%     \caption{ \textbf{Length-agnostic inference without retraining.} Each row corresponds to a distinct video sequence synthesized at different inference lengths (\eg, 49 denotes the total frame).  Both trained on fixed 81-frame clips, our model maintains high performance. In contrast, the direct concatenation method (\eg, ReCamMaster~\cite{ICCV_recammaster}) exhibits severe degradation when the inference horizon deviates from the training length, underscoring the brittleness of prefix-style conditioning.}
%     \label{fig:recam_vs_ours}
% \end{wrapfigure}

\begin{figure}
    \centering
    \includegraphics[width=0.95\linewidth]{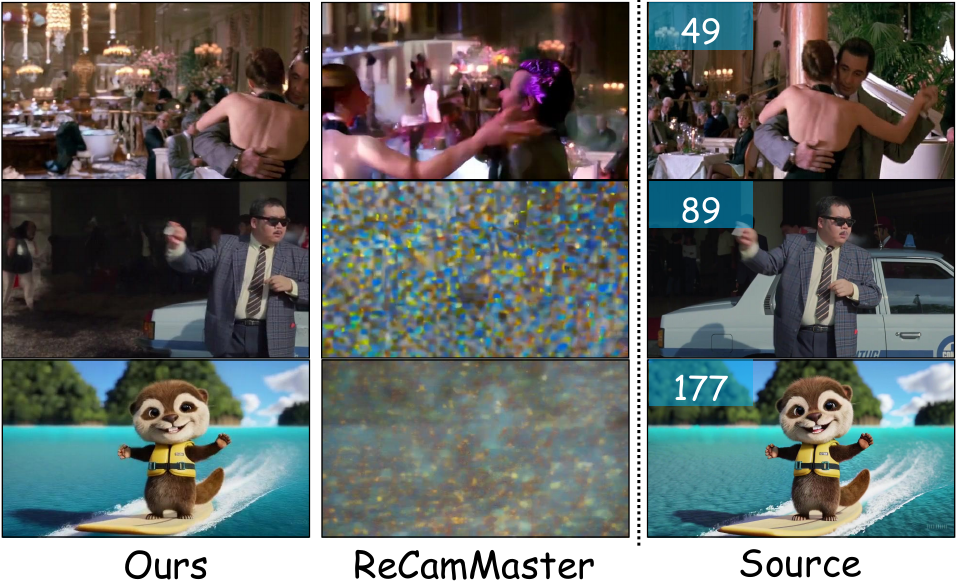}
    \vspace{-.6em}
    \caption{ \textbf{Length-agnostic inference without retraining.} Each row corresponds to a distinct video sequence synthesized at different inference lengths (\eg, 49 denotes the total frames of the generated video).  Both trained on fixed 81-frame clips, our model maintains high performance. In contrast, the direct concatenation method (\eg, ReCamMaster~\cite{ICCV_recammaster}) exhibits severe degradation when the inference horizon deviates from the training length, underscoring the brittleness of prefix-style conditioning.}
    \label{fig:recam_vs_ours}
\end{figure}

where $\texttt{Interleave}(z_{s}, z_t) \in \mathbb{R}^{b \times 2f \times s \times d}$ is the input of velocity prediction network,  $z_t$ is the noised $z_0$, $z_{s}^{i}$ is the i-th latent frame.
A pivotal advantage of this cross-frame design is the shift from absolute to relative modeling. By interleaving $z_s^i$ with its corresponding $z_t^i$ at each temporal position, the model learns to capture the relative frame relationships rather than being tethered to absolute temporal indices. This formulation enables the teacher model to generalize to arbitrary video lengths during inference without the need for retraining (as shown in Figure ~\ref{fig:recam_vs_ours}) and is compatible with the following causal adaptation described in Sec.~\ref{casual_adaption}. 
Like Recamaster~\cite{ICCV_recammaster}, we do not introduce additional parameters for cross-frame conditioning. And for camera condition injection, we first project the flattened 3 × 4 camera matrix to the space of video tokens and add it to the self-attention input features.

\textbf{Training Objective.}
To further capture the camera dynamics while preventing learning the static characteristics of the synthetic dataset, we introduce a motion-aware loss. 
This objective emphasizes temporal variations while suppressing redundant static appearance information.
Specifically, we compute the  frame-wise velocity differences 
$\bigtriangleup(v_\theta(\cdot))$ and $\bigtriangleup(v_t) \in \mathbb{R}^{(f-1)\times h\times w\times c}$, and we align them to preserve dynamic motion cues.
We define the motion loss as:
% \begin{equation}
% \label{eq:motion_loss}
%  \mathcal{L}_{Motion}  = \mathbb{E}_{z_0,z_1,t,c_{text},z_s,c_{cam}}\left[  \left \| \bigtriangleup (v_\theta(z_t,t,z_s,c_{text},c_{cam})) - \bigtriangleup (z_1 - z_0) \right \|^{2}_{2} \right].
% \end{equation}
\begin{equation}
\label{eq:motion_loss}
\begin{split}
    \mathcal{L}_{\text{Motion}} = \mathbb{E}_{z_0,z_1,t,c_{\text{text}},z_s,c_{\text{cam}}} \biggl[ & \Bigl\| \bigtriangleup \bigl( v_\theta(z_t,t,z_s,c_{\text{text}},c_{\text{cam}}) \bigr) \\
    & - \bigtriangleup (z_1 - z_0) \Bigr\|_2^2 \biggr].
\end{split}
\end{equation}
Then the total loss is the combination of the standard flow matching loss (Eq.~\eqref{eq:fm_training_objective}) on camera-conditioned video pairs and the motion loss as $\mathcal{L}_{Teacher} =( 1-\alpha) \cdot\mathcal{L}_{FM} + \alpha\cdot \mathcal{L}_{Motion}$, where $\alpha$ is the weighting coefficient.

% \textcolor{red}{and we set $\alpha = 0.15$ in practice}.

\subsection{Causal Distillation}
\label{train_student}
While the bidirectional teacher achieves high visual fidelity and camera controllability, its non-causal attention and multi-step attribute prevent real-time streaming.  We distill the teacher into a few-step autoregressive student model. Given a noise schedule $\mathcal{T} =\{t_0=0, \dots, t_N=T \}$, each frame is denoised over $N$ steps, where $N$ is significantly smaller than that in the multi-step teacher model, enabling real-time interactive control.

\textbf{Causal Adaptation.}
As shown in  Figure~\ref{fig:token_merge_strategy}
\label{casual_adaption}, our teacher model's interleaved structure naturally extends to the causal adaptation.
Following the design protocol from CausVid~\cite{yin2025Causvid}, we replace the bidirectional attention with causal attention architecture. And we pack the source and target noisy latent frame as a chunk $Z^i = [z_s^i, z_t^i]_{\text{frame-dim}} \in \mathbb{R}^{2\times h\times w\times c}$ as the input\footnote{In practice, generation is typically performed in chunks rather than one chunk.} of the causal model.
Previous works~\cite{yin2025Causvid,self_forcing} initilize the causal student using regression on ODE solution pairs sampled from the teacher. However, the sampling process is time-consuming. We directly initialize our causal student using the flow matching loss (Eq.~\eqref{eq:fm_training_objective}) on camera-conditioned video pairs.

\textbf{Distilling to Real-Time AR model.}
Following the previous works \cite{self_forcing, zhao2025real_motion}, we use Self-Forcing-style Distribution Matching Distillation framework to optimize our causal model. This approach explicitly simulates the autoregressive inference dynamics during training, effectively mitigating the exposure bias of the causal model after adaptation. 

During training, given a source video latent $z_s$ and pure noise $z_T \sim \mathcal{N}(0, \mathbf{I})$ , we partition them into $L$ chunks $\{z_s^i\}^L$ and $\{z_{t_j}^i\}_{j=N}^L$, where $t_j \in \mathcal{T}$. As shown in Figure~\ref{fig_pipe} (c), we interleave the source frames $z_s$ and noise frames $z_T$ as $\{Z_{t_N}^i\}^L$, where $Z_{t_N}^i = [z_s^i, z_{t_N}^i]$ is the i-th input chunk of the causal model. The sampling process of the i-th chunk
involves N-step iterative denoising on its input.
Specially, we randomly sample a denoising step $n$, denoise step-by-step from $Z_{t_N}^i$ to  $Z_{t_n}^i$. At each denoising step $t_j$, we get clean chunk $\hat{Z_{t_0}^i} = [\hat{z}_s^i, \hat{z}_{t_0}^i] $ by denoising the intermediate noisy frame $Z_{t_j}^i$ conditioned on a dynamically updated KV cache: $\mathcal{C}_i = \{Z_{t_j}^i\} \cup \{\hat{{Z}}^k_{t_0}\}_{\max(1, i-W) \leq k < i}$, where $\hat{Z}^k_{t_0}$ is the previously generated clean frames, $W$ is the attention window size of KV cache. We only preserve the target frame $\hat{z}_{t_0}^i$ and then injects Gaussian noise with a lower noise level into the predicted denoised clean frame via the forward diffusion process. This produces a less noisy frame $z_{t_{j-1}}^i$ which concate with the source frame $z_{s}^i$ as $Z_{t_{j-1}}^i$ as the input to the next denoising step.  
Formally, the denoising process is formulated as: $z_{t_{j-1}}^i =\Psi(G_{\theta}(Z_{t_j}^i,t_j, Z_0^{< i}),t_{j-1})$. 

Unlike Self-Forcing~\cite{self_forcing} updating 
the KV cache using the one-step noisy feature denoised from $Z_{t_{n}}^i$ to $\hat{Z}_{t_0}^i$,
we adopt the self-rollout strategy~\cite{zhao2025real_motion}, we continue denoising from $Z_{t_n}^i$ to $Z_{t_0}^i$ step-by-step, and  finally update the KV cache with the synthesized chunk $Z_{t_0}^i=[z_s^i, z_{t_0}^i]$.
Note the camera condition is also split into L chunks $\{c_{cam}^i\}^L$ to match the input of the causal student.
After completing the $L$ chunks in Self-Forcing manner, we obtain a fully generated video ${z}_0 = \{{z}^1_0, \dots, {z}^L_0\}$. We then apply the DMD objective to align the student's rollout distribution $p^{gen}$ with the teacher's data distribution $p^{\text{data}}$. The Eq.~\eqref{dmd_score} is reformulated as:

% \begin{equation}
% \label{eq:dmd-}
% \begin{split}
%     \nabla_\theta \mathcal{L}_{\text{DMD}} \approx -\mathbb{E}_{t, {z}_0, c_{\text{text}}, z_s, c_{\text{cam}}} \biggl[ & \Bigl( s_{\text{real}}(\Psi({z}_0, t), t, z_s, c_{\text{text}}, c_{\text{cam}})    \\ 
%     & - s_{\text{fake}}(\Psi({z}_0, t), t, z_s, c_{\text{text}}, c_{\text{cam}}) \Bigr) \cdot \frac{\partial {z}_0}{\partial \theta}  \biggr ].
% \end{split}
% \end{equation}

\begin{align}
\label{eq:dmd-}
    \nabla_\theta \mathcal{L}_{\text{DMD}} \approx -\mathbb{E}_{t, {z}_0, c_{\text{text}}, z_s, c_{\text{cam}}} \biggl[ & \Bigl( s_{\text{real}}(\Psi({z}_0, t), t, z_s, c_{\text{text}}, c_{\text{cam}})    \\ 
    & - s_{\text{fake}}(\Psi({z}_0, t), t, z_s, c_{\text{text}}, c_{\text{cam}}) \Bigr) \cdot \frac{\partial {z}_0}{\partial \theta}  \biggr ]. \notag
\end{align}

\textbf{Loop-Closed Data Augmentation (LoopAug).}
A critical issue we identify in causal distillation is \textit{loop inconsistency}: when the camera returns to its initial viewpoint in a closed-loop trajectory, the generated frame often exhibits visual inconsistency with the source video. 
We trace this to the absence of loop-consistent supervision in existing multi-view datasets. 
To address this without costly manual data collection, we introduce a data augmentation paradigm that synthesizes loop-closed sequences from existing multi-view datasets.  As demonstrated in Figure ~\ref{fig_pipe}(d), let $V = \{v^1,v^2,\dots,v^F \}$  
denote a raw video sequence of F frames. We construct a closed-loop extended video $V_e$ by concatenating the original sequence with its reversed counterpart $V_r = \{v^{F-1},v^{F-2}, \dots, v^1\}$: $V_e = [V,V_r]_{frame-dim} \in \mathbb{R}^{(2F-1)\times C \times H \times W}$. The corresponding camera condition follows the same process. This provides explicit supervision for loop consistency, encouraging the model to generate identical frames when the camera returns to the same viewpoint. In practice, we use this method to finetune the teacher model and causal adaptation with a truncation strategy.

%% file: tex/4-experiments.tex
\section{Experiments}
\label{sec:exp}
\begin{figure*}[t]
    \centering
    \includegraphics[width=1\textwidth]{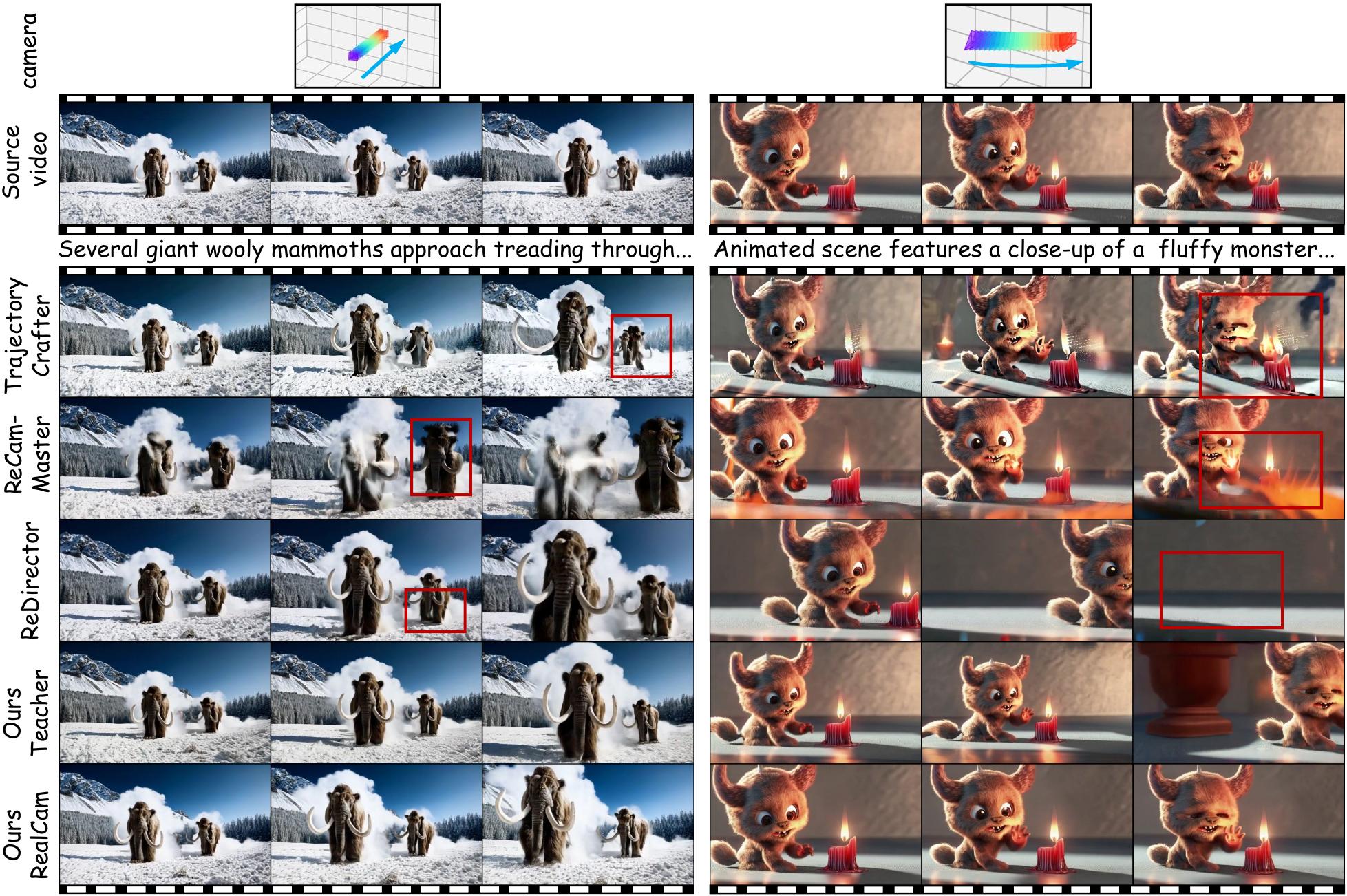}
    \vspace{-2em}
    \caption{\textbf{Qualitative comparison with SOTA methods.} \textcolor{red}{Red boxes} indicated low quality content across frames. Our method achieves better camera control and excellent temporal synchronization. }
    \vspace{-1em}
    \label{fig:qualitative_comparison}
\end{figure*}
\input{tabel/short_video_main}

\paragraph{Implementation Details.}
We build upon text-to-video (T2V) of Wan 2.1 (1.3B)~\cite{wang2025wan} and Wan 2.2 (5B)~\cite{wang2025wan}. We train the teacher model on the MultiCamVideo dataset~\cite{ICCV_recammaster} at a resolution of $81\times480 \times832$ with a learning rate of $0.0001$ and a batch size of 8. For causal adaptation and self-forcing distillation, we use a 3-step diffusion process in a chunk-wise manner containing 3 latent frames. We maintain a KV cache composed of a chunk from the initial frames and a fixed-size local window of the recent 5 chunks. We use LoopAug with a truncation strategy to further finetune the teacher model and train the causal student. All training is performed using the AdamW optimizer~\cite{adam} on NVIDIA H20 GPUs. We refer readers to the Appendix for additional details.

\textbf{Baselines.}
We compare our method with the state-of-the-art
camera-controlled V2V generation models~\cite{CVPR_redirector, ICCV_TrajectoryCrafter, ICCV_recammaster}. We adopt each baseline’s official backbone and implementation, allowing them to fully demonstrate designed capabilities.
TrajectoryCrafter~\cite{ICCV_TrajectoryCrafter} is the explicit warping-based method, which leverages a monocular depth model to preprocess the input video.  ReCamMaster~\cite{ICCV_recammaster} and Redirector~\cite{CVPR_redirector} are implicit pose-based methods conditioned on camera extrinsic parameters.  Besides, Redirector needs to use an external model~\cite{huang2025vipe} to predict the input video's camera extrinsic parameters and intrinsic parameters.

\textbf{Evaluation Protocol.}
To ensure a rigorous and comprehensive evaluation, we construct a diverse test set consisting of both short and long video sequences. The short-video set comprises 30 81-frame videos, including 15 real-world samples from~\cite{ICCV_recammaster} and 15 synthetic samples from the Sora webpage~\cite{brooks-sora}. To evaluate long-horizon performance, we curate an additional 20 177-frame videos (5 real-world and 15 synthetic). For each source video, we apply 10 distinct camera trajectories following the protocol in ReCamMaster~\cite{ICCV_recammaster}, yielding a total of 300 (short) and 200 (long) unique test cases. 

We evaluate the generated videos across three key dimensions: 1) Visual Quality: We employ multiple metrics from the VBench suite~\cite{huang2024vbench} to assess Aesthetic Quality (Aes. Qual.), Imaging Quality (Img. Qual.), Temporal Flickering (Tem. Flick.), Motion Smoothness (Motion Smooth.), Subject Consistency (Sub. Cons.), and Background Consistency (Bg. Cons.). 2) Geometric Consistency: Following~\cite{CVPR_redirector}, we use Dyn-MEt3R~\cite{Dyn-MEt3R} to measure the global geometric consistency of the generated videos, and per-frame MEt3R~\cite{asim2025met3r} to quantify structural alignment with the source video. 3) Camera Accuracy: We report TransErr and RotErr~\cite{ICCV_recammaster, CVPR_redirector}, which measure the relative translation and rotation errors for every frame pair. The camera poses of the generated videos are estimated using ViPE~\cite{huang2025vipe} for ground-truth comparison. Finally, we report the first-frame latency measured on a single NVIDIA H20 GPU as the primary indicator of real-time interactive performance.

\subsection{Results}
\textbf{Qualitative Comparisons.}
We present qualitative comparisons between our RealCam and several state-of-the-art baselines in Figure~\ref{fig:qualitative_comparison}. As highlighted by the red boxes, existing methods frequently suffer from content distortion and temporal desynchronization. For instance, TrajectoryCrafter~\cite{ICCV_TrajectoryCrafter} introduces noticeable ghosting and structural artifacts when handling large camera displacements. Implicit methods like ReCamMaster~\cite{ICCV_recammaster} exhibit significant motion blur and fail to maintain the identity of dynamic subjects (\eg, the mammoths) as the sequence progresses. While ReDirector~\cite{CVPR_redirector} attempts to preserve geometry, it occasionally fails to localize dynamic objects under rapid viewpoint shifts, leading to the complete disappearance of foreground elements (as seen in the monster scene). In contrast, our Teacher model and its causal counterpart, RealCam, consistently synthesize high-fidelity videos that strictly adhere to the specified camera trajectories while maintaining seamless temporal synchronization with the source video. Notably, despite moving to an autoregressive streaming architecture, RealCam achieves visual quality and camera control accuracy comparable to the bidirectional Teacher model.

\textbf{Quantitative Comparisons.}
The overall performance comparisons are summarized in Table~\ref{tab:main_quanti}, yielding several key observations:  RealCam achieves a dramatic reduction in latency compared to existing bidirectional frameworks. While state-of-the-art methods like TrajectoryCrafter and ReDirector require hundreds of seconds to process a single clip, our causal models (1.3b and 5b) achieve sub-second inference speeds (\eg, 0.72s for the 5b variant), representing an orders-of-magnitude speedup that is essential for interactive applications.

% \begin{wraptable}[12]{r}
% {0.6\textwidth}
%  \vspace{-1.2em}
%     \centering
%     \caption{\textbf{User study results.} We evaluate video quality and camera following capability through pairwise comparisons. Our method is preferred (over $>$ 50\%) over most baselines, with the teacher being slightly preferred over the student.} 
%     \small 
%     \vspace{-0.8em}
%     \label{tab:user}
%     \begin{tabular}{l c c c c}
%     \toprule
%          & \multicolumn{2}{c}{Video Quality} & \multicolumn{2}{c}{Camera Following} \\
%     \cmidrule(lr){2-3} \cmidrule(lr){4-5}
%         Method & Ours-T & Ours-C & Ours-T & Ours-C \\
%     \midrule
%         TrajectoryCrafter~\cite{ICCV_TrajectoryCrafter} & 66.34\% & 59.77\% & 64.62\% & 58.81\% \\
%         ReCamMaster~\citep{ICCV_recammaster} & 53.60\% & 51.45\% & 54.25\% & 51.00\% \\
%         ReDirector~\cite{CVPR_redirector} & 51.24\% & 50.52\% & 50.85\% & 49.37\% \\
%     \midrule
%          Ours-C & 50.65\% & - & 51.1\% & - \\
%     \bottomrule
%     \end{tabular}
% \end{wraptable}

Despite the transition to a streaming-capable, autoregressive architecture, 
RealCam demonstrates superior visual quality and consistency. Notably, our models outperform all baselines in visual quality, which we attribute to the Cross-frame In-context Learning paradigm that effectively anchors the target frames to the source context. In terms of geometric consistency, our models attain the highest Dyn-MEt3R scores and lowest MEt3R scores, 
highlighting their strength in preserving 3D structure during dynamic camera movement. Furthermore, the camera accuracy metrics remain competitive with the teacher model and baselines, particularly in Trans
Err. It is worth noting that the performance gap between our Teacher and Causal student is minimal across most metrics, validating that our self-rollout distillation process successfully preserves the teacher’s high-fidelity geometric reasoning while enabling efficient, on-the-fly synthesis.

% \begin{wraptable}{r}{0.65\textwidth} 
% \centering
% \input{tabel/user_study}
% \end{wraptable}

\textbf{User Study.}
We further perform human evaluation to reflect real preferences. We collected 1,000 responses evaluating video quality and camera following capability of generated videos using our Wan 2.1 (1.3B) variants. Since accurately assessing various metrics is challenging for participants, we focus on two key dimensions: visual quality and camera following. As shown in Table~\ref{tab:user}, our teacher model (Ours-T) consistently outperforms all baselines in both dimensions. Our causal student (Ours-C) maintains superior video quality over all baselines and achieves better camera control, with the only exception being ReDirector~\cite{CVPR_redirector}.
 Notably, ReDirector relies on an external model~\cite{huang2025vipe} to estimate camera parameters from the source video, which provides it with an advantage in camera alignment tasks.
\begin{table}%
\caption{\textbf{User study results.} We evaluate video quality and camera following capability through pairwise comparisons. Our method is preferred (over $>$ 50\%) over most baselines, with the teacher being slightly preferred over the student.} 
\vspace{\belowtabcapmargin}
\label{tab:user}
\begin{minipage}{\columnwidth}
\begin{center}
   \small 
   \setlength\tabcolsep{3pt}
    \begin{tabular}{l c c c c}
    \toprule
         & \multicolumn{2}{c}{Video Quality} & \multicolumn{2}{c}{Camera Following} \\
    \cmidrule(lr){2-3} \cmidrule(lr){4-5}
        Method & Ours-T & Ours-C & Ours-T & Ours-C \\
    \midrule
        TrajectoryCrafter~\cite{ICCV_TrajectoryCrafter} & 66.34\% & 59.77\% & 64.62\% & 58.81\% \\
        ReCamMaster~\citep{ICCV_recammaster} & 53.60\% & 51.45\% & 54.25\% & 51.00\% \\
        ReDirector~\cite{CVPR_redirector} & 51.24\% & 50.52\% & 50.85\% & 49.37\% \\
    \midrule
         Ours-C & 50.65\% & - & 51.1\% & - \\
    \bottomrule
    \end{tabular}
\end{center}
 
\end{minipage}
\end{table}%

\begin{figure*}[t]
    \centering
    \includegraphics[width=1\textwidth]{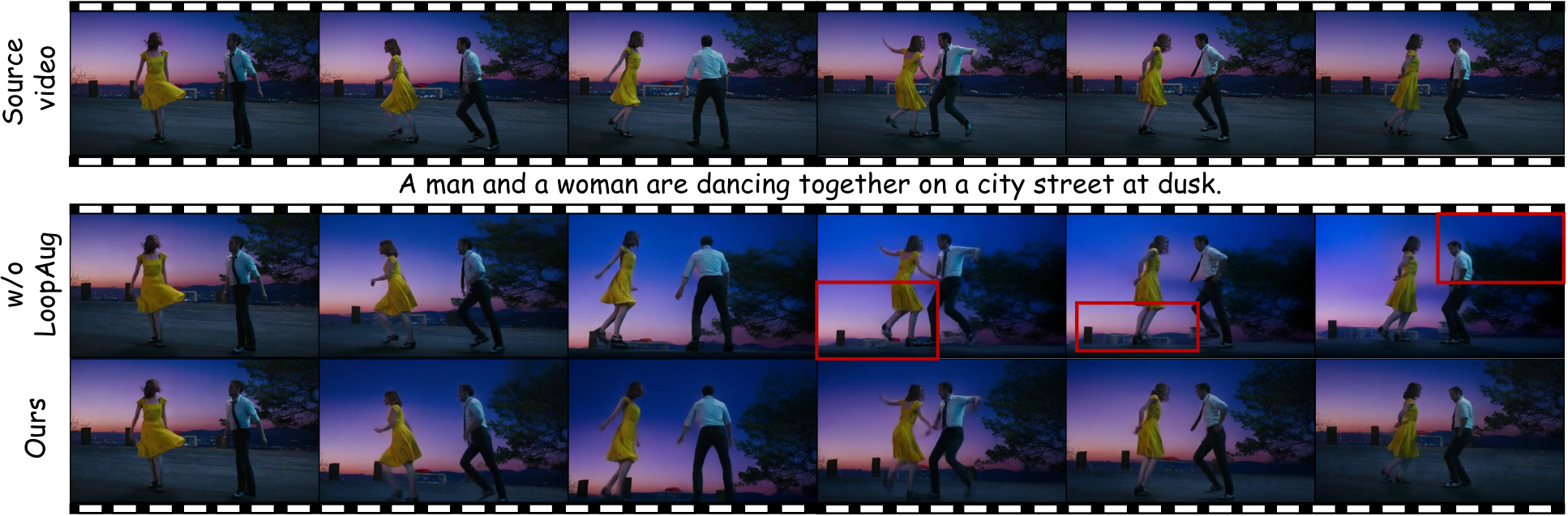}
    \vspace{-2em}
    \caption{\textbf{Qualitative ablation on long video.} The camera trajectory first translates down with rotation and then back to the origin. \textcolor{red}{Red boxes} indicated inconsistency with the source video. }
    % \vspace{-1em}
    \label{fig:qualitative_ablation}
\end{figure*}

\subsection{Ablation Study}
\input{tabel/ablation-81}
\textbf{Impact of Chunk Size.}
We investigate the latent chunk size, a key design choice that governs the balance between streaming quality and interactivity. As shown in Table~\ref{tab:main_ablation}, while a chunk size of 1 minimizes first-frame latency ($0.42s$), it suffers from significant quality degradation and high camera following error, especially when the videos extend to  177 frames.  Increasing the chunk size to 3 provides the model with a broader field for bidirectional modeling, leading to a substantial boost in visual fidelity and a marked improvement in geometric consistency and camera accuracy. Given that a chunk size of 3 still maintains sub-second responsiveness ($0.72s$), we select it as our optimal configuration to ensure precise camera control without compromising real-time interactivity.

\textbf{Impact of Loop-Closed Data Augmentation.}
We further validate the effectiveness of LoopAug in enhancing long-term consistency and global stability. The impact of LoopAug is most prominent when scaling to long-horizon generation.  
Comparing the configurations with and without LoopAug in Table~\ref{tab:main_ablation}, the augmentation yields consistent improvements across all visual and geometric metrics.  On 177-frame sequences, our full model preserves structural integrity and camera alignment without performance drop-off, whereas the ablated variant exhibits loop inconsistency ( \textcolor{red}{Red boxes} in Figure~\ref{fig:qualitative_ablation}). This confirms that loop-closed supervision provides essential global constraints, enabling the causal student to maintain coherence over long video lengths.

%% file: tabel/short_video_main.tex
\begin{table*}[t!]
    \centering
    \caption{\textbf{Quantitative comparison with SOTA methods.} Our method improves visual quality and keeps excellent geometric consistency and camera control, while significantly reducing latency.}
    \vspace{\belowtabcapmargin}
    \label{tab:main_quanti}
    \setlength\tabcolsep{4pt}
    \resizebox{\linewidth}{!}{
    \begin{tabular}{lccccccccccc}
       \toprule
        \multirow{2}{*}[-1.33ex]{Method} & \multicolumn{1}{c}{Latency}  &\multicolumn{6}{c}{Visual quality$\uparrow$} & \multicolumn{2}{c}{Geometric Consistency} & \multicolumn{2}{c}{Camera Accuracy}  \\
        \arrayrulecolor{gray}\cmidrule(lr){3-8} \cmidrule(lr){9-10} \cmidrule(lr){11-12} & \makecell{(s) $\downarrow$}
        & \makecell{Sub. \\ Cons}  & \makecell{Bg. \\ Cons.} &  \makecell{Aes. \\ Qual.}  & \makecell{Img. \\ Qual.} &\makecell{Tem. \\ Flick.} & \makecell{Motion \\ Smooth}  & \makecell{Dyn \\ MEt3R}$\uparrow$ & MEt3R$\downarrow$ & \makecell{ Trans \\ Err}$\downarrow$ & \makecell{Rot\\Err}$\downarrow$  \\
        \midrule
        ReCamMaster~\cite{ICCV_recammaster} & 426 & 91.65 & 93.72 & 58.83 & 57.31 & \cellcolor{tabthird}97.12 & 99.08 & 0.7846 & 0.2864 & 0.0301 & 3.150 \\
        TrajectoryCrafter~\cite{ICCV_TrajectoryCrafter} &687   & 90.97 & 93.22 & 59.17 & \cellcolor{tabsecond}62.89 & 96.15 & 97.49 & 0.7610 & \cellcolor{tabsecond}0.2268 & 0.0705 & 5.107 \\
        ReDirector~\cite{CVPR_redirector} &613 & 91.81 & 93.77 & 57.22 & 58.59 & 96.60 & 99.11 & 0.8011 & 0.2786 & \cellcolor{tabfirst}0.0236 & \cellcolor{tabsecond}2.866 \\
        \arrayrulecolor{gray} \midrule
         \textbf{Ours Teacher (1.3b)} & 426& 92.16 & 94.06 &  59.46 & \cellcolor{tabthird}62.09 & 97.09 & \cellcolor{tabthird}99.21 & \cellcolor{tabthird}0.8248 & 0.2643 & \cellcolor{tabthird}0.0255 & \cellcolor{tabthird}2.874 \\
         \textbf{Ours Causal (1.3b)} & 1.15&\cellcolor{tabthird}92.61 & \cellcolor{tabsecond}94.73 & \cellcolor{tabthird}59.68 & \cellcolor{tabfirst}63.02 & \cellcolor{tabsecond}97.24 & \cellcolor{tabsecond}99.22 & 0.8136 & \cellcolor{tabfirst}0.2245 & 0.0300 & 3.312 \\
         \textbf{Ours Teacher (5b)} & 232&\cellcolor{tabfirst}93.19 & \cellcolor{tabthird}94.61 & \cellcolor{tabsecond}59.79 & 60.02 & 97.09 & 99.19 & \cellcolor{tabsecond}0.8275 & \cellcolor{tabthird}0.2639 & \cellcolor{tabsecond}0.0253 & \cellcolor{tabfirst}2.616 \\
         \textbf{Ours Causal (5b)} & 0.72& \cellcolor{tabsecond}93.01 & \cellcolor{tabfirst}94.97 & \cellcolor{tabfirst}60.03 & 59.98 & \cellcolor{tabfirst}97.31 & \cellcolor{tabfirst}99.23 & \cellcolor{tabfirst}0.8282 & 0.2650 & 0.0287 & 3.335 \\
       \arrayrulecolor{black}\bottomrule
    \end{tabular}
    }
    \vspace{\abovetabcapmargin}
\end{table*}

%% file: tabel/ablation-81.tex
\begin{table*}[t!]
    \centering
    \caption{\textbf{Quantitative ablations on key training strategies.} }
    \vspace{\belowtabcapmargin}
    \label{tab:main_ablation}
    \setlength\tabcolsep{4pt}
    \resizebox{\linewidth}{!}{
    \begin{tabular}{lccccccccccc}
       \toprule
        \multirow{2}{*}[-1.33ex]{Method} & \multicolumn{1}{c}{Latency}  &\multicolumn{6}{c}{Visual quality$\uparrow$} & \multicolumn{2}{c}{Geometric Consistency} & \multicolumn{2}{c}{Camera Accuracy}  \\
        \arrayrulecolor{gray}\cmidrule(lr){3-8} \cmidrule(lr){9-10} \cmidrule(lr){11-12} & \makecell{(s) $\downarrow$}
        & \makecell{Sub. \\ Cons}  & \makecell{Bg. \\ Cons.} &  \makecell{Aes. \\ Qual.}  & \makecell{Img. \\ Qual.} &\makecell{Tem. \\ Flick.} & \makecell{Motion \\ Smooth}  & \makecell{Dyn \\ MEt3R}$\uparrow$ & MEt3R$\downarrow$ & \makecell{ Trans \\ Err}$\downarrow$ & \makecell{Rot\\Err}$\downarrow$  \\
        \midrule
        Chunk=1 (w/o LoopAug) & \cellcolor{tabfirst}0.42 & 91.97 & 93.78 & 59.22 & 57.98 & 96.73 & 99.14 & 0.8200 & \cellcolor{tabfirst}0.2602 & 0.0308 & 4.5516 \\
        Chunk=3 (w/o LoopAug) & 0.72 & \cellcolor{tabfirst}93.14 & 94.63 & \cellcolor{tabfirst}60.20 & \cellcolor{tabfirst}60.12 & 96.91 & 99.17 & 0.8167 & 0.2634 & \cellcolor{tabfirst}0.0257 & \cellcolor{tabfirst}3.137 \\
         {Ours } & 0.72& 93.01 & \cellcolor{tabfirst}94.97 & 60.03 & 59.98 & \cellcolor{tabfirst}97.31 & \cellcolor{tabfirst}99.23 & \cellcolor{tabfirst}0.8282 & 0.2650 & 0.0287 & 3.335 \\
  % --- 新增的分隔线和标题 ---
        \arrayrulecolor{gray}\midrule
        \multicolumn{12}{c}{\textbf{results on long video (177) frames}} \\
        \midrule
        
        % --- 新增的数据部分 ---
        Chunk=1 (w/o LoopAug) & \cellcolor{tabfirst}0.42 & 89.70 & 92.65 & 58.75 & 59.47 & 97.01 & 99.15 & 0.7471 & 0.3041 & 0.1269 & 9.0987 \\
        Chunk=3 (w/o LoopAug) & 0.72 & 92.30 & 94.24 & 60.82 & 62.63 & 97.09 & 99.14 & 0.7905 & 0.2920  & 0.1238 & 6.5733 \\
        {Ours } & 0.72 & \cellcolor{tabfirst}93.89 & \cellcolor{tabfirst}95.09 & \cellcolor{tabfirst}61.82 & \cellcolor{tabfirst}64.15 & \cellcolor{tabfirst}97.11 & \cellcolor{tabfirst}99.20 & \cellcolor{tabfirst}0.7920  & \cellcolor{tabfirst}0.2644 & \cellcolor{tabfirst}0.0596 & \cellcolor{tabfirst}3.7419 \\
        
       \arrayrulecolor{black}\bottomrule
    \end{tabular}
    }
    \vspace{\abovetabcapmargin}
\end{table*}

%% file: tex/5-conclusion.tex
\section{Conclusion}
We present \texttt{RealCam}, a novel few-step autoregressive video diffusion framework that enables interactive, real-time camera-controlled V2V generation. By refactoring video conditioning into a Cross-frame In-context Learning paradigm, we overcome the structural bottlenecks of rigid prefix-style methods, achieving length-agnostic synthesis and seamless causal adaptation. Coupled with our LoopAug strategy, \texttt{RealCam} effectively suppresses long-horizon drift while delivering state-of-the-art visual fidelity at sub-second latency. Crucially, our causal student achieves camera control performance and visual quality competitive with the bidirectional teacher model, proving that real-time efficiency can be achieved without sacrificing geometric precision. Limitations and future directions are discussed in Appendix.